\documentclass{article}

\usepackage[final]{corl_2026} 

\makeatletter
\renewcommand{\@conferencelocation}{Austin, Texas, USA}
\makeatother

\usepackage{amsmath,amssymb}
\usepackage{booktabs}
\usepackage{graphicx}
\usepackage[table]{xcolor}
\usepackage{multirow}
\usepackage{hyperref}
\usepackage{capt-of}
\usepackage{pifont}
\usepackage{wrapfig}
\usepackage{microtype}
\usepackage{tikz}
\usetikzlibrary{shadows}

\newcounter{algobox}
\newcounter{algoline}
\newcounter{algoindent}
\newlength{\algoindentwidth}
\newcommand{\algolinefmt}[1]{%
  \stepcounter{algoline}%
  \noindent\makebox[1.8em][r]{\thealgoline:}\hspace{0.4em}%
  \hspace*{\dimexpr\value{algoindent}\algoindentwidth\relax}#1\par
}
\newenvironment{algoic}{%
  \setcounter{algoline}{0}%
  \setcounter{algoindent}{0}%
  \setlength{\parskip}{0pt}%
}{}
\newcommand{\Require}[1]{\algolinefmt{\textbf{Require} #1}}
\newcommand{\State}[1]{\algolinefmt{#1}}
\newcommand{\If}[1]{%
  \algolinefmt{\textbf{if} #1 \textbf{then}}%
  \addtocounter{algoindent}{1}%
}
\newcommand{\ElsIf}[1]{%
  \addtocounter{algoindent}{-1}%
  \algolinefmt{\textbf{else if} #1 \textbf{then}}%
  \addtocounter{algoindent}{1}%
}
\newcommand{\Else}{%
  \addtocounter{algoindent}{-1}%
  \algolinefmt{\textbf{else}}%
  \addtocounter{algoindent}{1}%
}
\newcommand{\EndIf}{%
  \addtocounter{algoindent}{-1}%
  \algolinefmt{\textbf{end if}}%
}


\title{What is the Better Curriculum:\\
Controller-Shaped Grasping Behavior\\
for Contact Force-Sensitive Manipulation}

\author{%
\textbf{Ziyan Feng} \quad
\textbf{Zizhao Yuan} \quad
\textbf{Yulong Fu} \quad
\textbf{Yuxin He}\\
\textbf{Zhiyuan Zhang} \quad
\textbf{Zhengjie Zhang} \quad
\textbf{Jinni Zhou} \quad
\textbf{Renjing Xu} \quad
\textbf{Qiang Nie}\begin{NoHyper}\thanks{Corresponding author: \texttt{qiangnie@hkust-gz.edu.cn}.}\end{NoHyper}\\
\textnormal{The Hong Kong University of Science and Technology (Guangzhou)}
}

\hypersetup{
  hyperfootnotes=false,
  pdftitle={What is the Better Curriculum: Controller-Shaped Grasping Behavior for Contact Force-Sensitive Manipulation},
  pdfauthor={Ziyan Feng, Zizhao Yuan, Yulong Fu, Yuxin He, Zhiyuan Zhang, Zhengjie Zhang, Jinni Zhou, Renjing Xu, Qiang Nie},
  pdfkeywords={Controller-Shaped Grasping Behavior, Vision-Based Tactile Sensing, Demonstration Shaping, Policy Learning, Force-Sensitive Manipulation}
}

\makeatletter
\g@addto@macro\@maketitle{%
  \begin{center}
    \includegraphics[width=\linewidth]{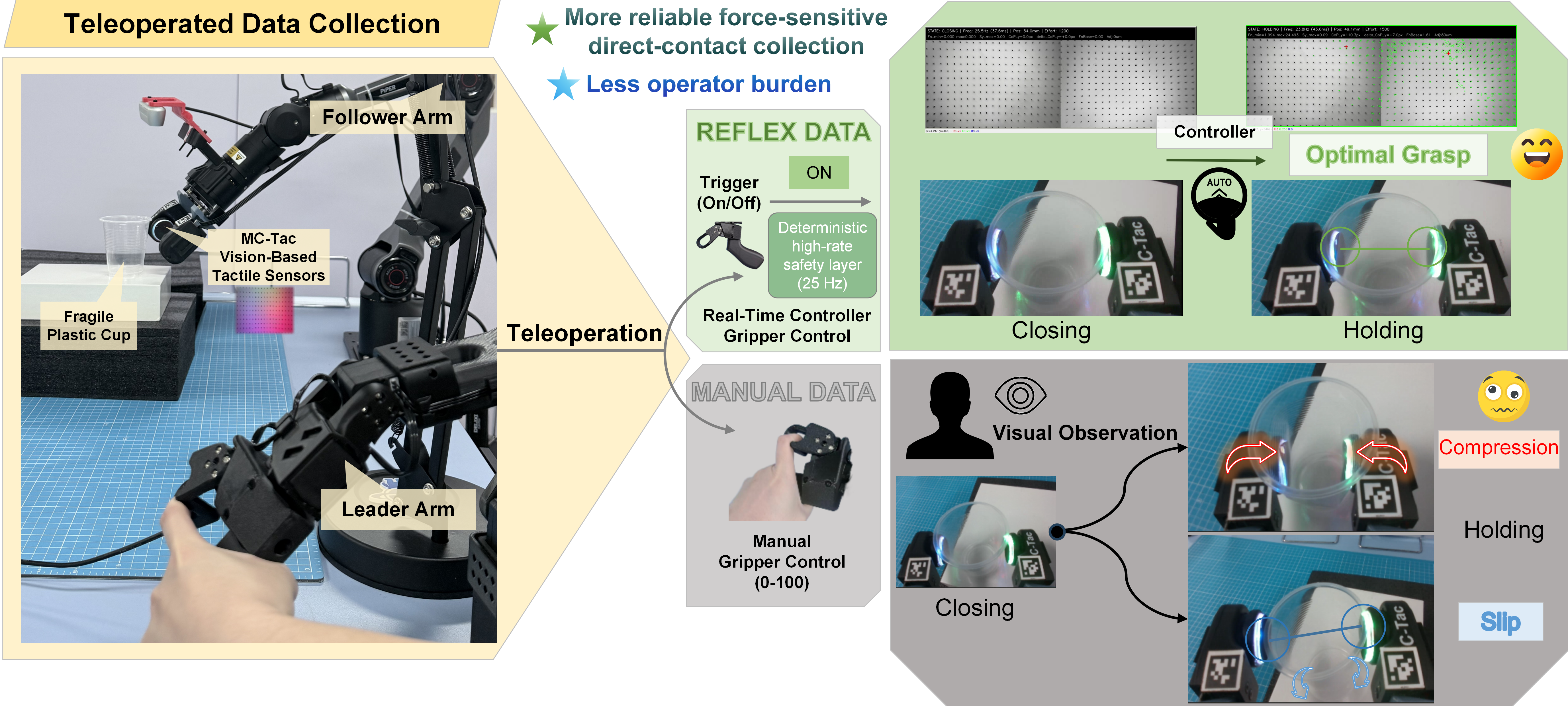}
    \captionof{figure}{Teleoperated fragile-cup collection. A leader arm
    controls the follower arm. Both modes provide on-site views and a
    graphical user interface displaying live tactile parameters and slip
    events. In manual mode,
    these cues are advisory and the operator uses coarse gripper buttons;
    only reflex mode uses TactileReflex~\citep{tactilereflex2025} to close
    a direct 25\,Hz tactile--action loop.}
    \label{fig:teaser}
  \end{center}
  
}
\makeatother

\begin{document}
\maketitle

\begin{abstract}
How should a robot learn to manipulate objects so fragile that sub-Newton
contact forces can cause irreversible damage? Existing visuo-tactile policy
learning typically treats tactile sensing as an additional policy input. In
direct-contact force-sensitive manipulation, however, the bottleneck can arise
earlier, during data collection: manual gripper control is too delayed and
coarse-grained to reliably maintain the narrow force range required for stable
grasping. We
therefore use a deterministic 25\,Hz tactile reflex controller as a
collection-time teacher, producing demonstrations with controller-shaped
grasping behavior for tactile-free policy learning. On Action Chunking with
Transformers (ACT), policies trained from reflex-shaped demonstrations recover
the teacher's grasping profile and achieve 95\% stable grasps on the nominal
plastic-cup task, substantially outperforming visually screened manual
demonstrations. The same intervention improves in-distribution stability on
$\pi_{0.5}$ and shows a favorable exploratory trend on an unseen paper-cup
variant. Under randomized external disturbance, however, the reflex-data
$\pi_{0.5}$ policy still fails in 45\% of policy-only trials, whereas a
deployment-time reflex arbiter retains all grasps. These results reveal a
new role for tactile feedback in force-sensitive manipulation: rather than
integrating tactile into the policy, we use it as a collection-time teacher that
shapes grasping behavior in demonstrations for policy learning, while
disturbance rejection remains controller-dependent, revealing the boundary of
tactile-free policy.
\par\smallskip\noindent
\mbox{Project website: \url{https://shayfeng.github.io/better-curriculum/}.}
\end{abstract}


\section{Introduction}
\label{sec:intro}

Imagine lifting a disposable plastic cup: 3.5\,g, 0.3\,mm wall
thickness. Gripping too lightly leads to slip, while exceeding the
viable force range by a few hundred millinewtons causes irreversible
deformation. Classical tactile
controllers handle such objects~\citep{romeo2020,weiner2021grasp,forte2025}
through deterministic high-rate feedback, but they lack the multi-step
task capability of learned policies.
Learned policies such as Action Chunking with Transformers
(ACT)~\citep{zhao2023act}, diffusion-policy-style
models~\citep{chi2023diffusion}, and $\pi_{0.5}$~\citep{pi05_2025}
support multi-step task execution but do not provide closed-loop 
force modulation.
Recent visuo-tactile policies~(Sec.~\ref{sec:related}) improve
contact-rich manipulation, yet in direct-contact force-sensitive container
manipulation the main difficulty begins earlier: manual gripper control during
teleoperated data collection relies on a human
observation--response--execution loop, whose cumulative perception, reaction,
and actuation delays can be too large for timely correction when the gripper
must remain within an extremely narrow contact-force range. In our collection
setup, the operator is given on-site visual access together with a graphical
user interface (GUI) displaying live tactile parameters and slip events,
allowing the grasp state to be monitored online. Even with these real-time
cues, however, the operator must still interpret the contact state and convert
it into coarse gripper commands through the same delayed human loop. Unlike
many contact-rich tasks where contact is mediated by insertion geometry or tool
interaction, direct-contact fragile grasping requires the gripper itself to
continuously regulate this narrow force window on the object surface throughout
collection.

This bottleneck is easy to miss if tactile is treated primarily as an
additional modality to integrate into policy learning or execution.
Classical tactile control stabilizes contact after grasp 
formation~\citep{veiga2020grip,weiner2021grasp,forte2025}, whereas recent visuo-tactile 
policy learning studies how the tactile modality can be integrated into the 
learned policy~\citep{calandra2018,vitac3d2024,kinedex2025}. 
Neither route isolates the upstream
learning problem here: before policy optimization can succeed, good
demonstrations must first be made repeatable enough to collect.

In this work, we address this limitation with a deterministic
vision-based tactile reflex controller used as a collection-time
teacher. Rather than integrating tactile into the policy, we run
TactileReflex~\citep{tactilereflex2025} at 25\,Hz during teleoperated data
collection to autonomously regulate follower-gripper force in real
time. This produces demonstrations with controller-shaped grasping behavior
for tactile-free policy learning while keeping the policy training
and deployment interfaces unchanged. The logged actions record the follower's
controller-regulated realized trajectories rather than unreconciled leader
commands, and a tactile-free student later learns to reproduce that behavior
at inference. This makes the
approach easy to pair with larger policy backbones without requiring
tactile-specific policy changes. We use ``curriculum'' as shorthand
for this controller-shaped demonstration-data distribution, not for a
staged sample schedule.
Fig.~\ref{fig:teaser} previews this collection-time mechanism and the
resulting demonstration-quality difference, and
Fig.~\ref{fig:architecture} formalizes deployment-time decoupling.

This insertion point is well suited to fragile contact
manipulation because the central difficulty is not only learning a
control law, but making successful contact behavior collectable and
therefore learnable. Without reflex assistance, operators use on-site
scene observations and live tactile parameters and slip events displayed
in the collection GUI, but must convert these visual cues into delayed,
coarse button commands. Grasps may therefore briefly enter the right
regime but are hard to maintain repeatably under a fixed manual protocol.
With the reflex active, that same narrow regime becomes repeatably
realizable during collection. We use a tactile-free student to test
whether tactile improves learning by shaping demonstrations rather
than by expanding the learned policy interface. More broadly, tactile
sensing remains less standardized and data-rich than red-green-blue (RGB)
vision, spanning
pressure maps, force/torque signals, binary contact indicators, and
high-resolution tactile images, often with comparatively small or
closed datasets~\citep{cao2026tactilefusion}. In the current setup, naive
tactile-image fine-tuning did not improve over the vision-only route
(Appendix~\ref{app:act_tactile}), so the core question here is the
collection-time role of tactile rather than richer visuo-tactile
policy integration. For the tactile-free student itself, the controller's
tactile regulation is transferred only through these demonstrations.

We evaluate this central claim through three main lines of evidence. On ACT, controller-shaped demonstrations provide the clearest
controlled evidence that this data intervention helps a tactile-free policy recover the nominal contact
regime; on $\pi_{0.5}$, the same data intervention improves in-distribution (ID) stability, showing that the
data-source effect is not ACT-specific, and yields favorable exploratory out-of-distribution (OOD)
evidence on an unseen paper-cup variant. Under randomized external disturbance, adding a high-rate
deterministic arbiter enables zero-failure grasp retention. Taken together, these results support our
central claim: controller-shaped demonstrations provide a more consistent training distribution for
tactile-free policy learning, while high-rate tactile feedback complements the learned policy by providing
rapid disturbance rejection at deployment.

\textbf{Contributions.}\quad

\textbf{\ding{182} Controller-shaped demonstrations as a training-data intervention.}
We identify a new collection-time role for tactile sensing in fragile contact manipulation:
a deterministic vision-based tactile reflex controller acts as a teacher that produces
teleoperated demonstrations with controller-shaped grasping behavior. Together, these
shaped demonstrations provide a more consistent training distribution for learning tactile-free
policies in an otherwise data-scarce and precision-critical grasp regime.

\textbf{\ding{183} Learning controller-shaped grasping behavior with a tactile-free policy.}
We show that controller-shaped demonstrations improve tactile-free
policy learning of nominal grasp behavior. Across ACT and $\pi_{0.5}$,
policies trained on reflex-shaped demonstrations achieve substantially
better in-distribution grasp quality than visually screened manual-data
baselines, without tactile-specific changes to the policy
training or deployment interface.

\textbf{\ding{184} Revealing the boundary of tactile-free policy.}
We introduce a deployment-time disturbance protocol that separates
nominal regime learning from fast tactile reactivity. Under randomized
external perturbations, the same policy requires a
decoupled high-rate gripper arbiter for reliable grasp retention,
revealing the boundary of what collection-time tactile shaping can
implicitly transfer.


\section{Related Work}
\label{sec:related}

\textbf{Visuo-tactile policy learning.}\quad
Recent visuo-tactile manipulation
policies~\citep{calandra2018,vitac3d2024,mimictouch2024,kinedex2025,vtla2025,vlatouch2025,tactilevla2026,omnivtla2025,dreamtacvla2025,tafvla2026,tacvla2026}
often study insertion, tool use, and other contact-rich 
tasks by treating tactile as an additional modality in policy learning
or execution. Crucially, they typically do not ask whether tactile can
improve learning upstream by making low-yield force-sensitive
collection repeatable enough to train from. Recent CoRL work also
studies dexterous teleoperation and collection interfaces for skill
transfer, such as DexUMI~\citep{dexumi2025} and
TypeTele~\citep{typetele2025}, but these works focus on embodiment gap
reduction, retargeting, or operator interfaces rather than isolating
collection-time tactile shaping for direct fragile-object
manipulation. A concurrent work, HapticVLA~\citep{hapticvla_2026},
removes inference-time tactile sensing by distilling a tactile-aware
token from a reward-weighted action expert. By contrast, a
deterministic reflex controller directly shapes collection, yielding
controller-shaped grasping demonstrations for tactile-free policy
learning.

\textbf{Privileged information distillation.}\quad
Learning from privileged information~\citep{vapnik2015} is well
established in robotics teacher-student
frameworks~\citep{chen2020cheating,lee2020learning,kumar2021rma,agarwal2023legged},
where the privileged teacher is typically a \emph{learned} policy or a
richer observation stream. Here, the teacher is an \emph{engineered,
deterministic} controller active only during collection. It does not
provide an extra modality to the learned policy; instead, it writes a
more stable nominal contact regime into the demonstrations. Here,
``distillation'' is interpretive and limited to collection time: the
student reproduces controller-shaped grasping behavior without
inheriting full tactile reactivity or using tactile input at inference.

\textbf{Reactive control and deployment-time safety.}\quad
RDP and ImplicitRDP~\citep{rdp2025,implicitrdp2025}, CompliantVLA~\citep{compliantvla2026},
VLSA~\citep{vlsa2025}, SafeDiff~\citep{safediff2025}, and
Safe-Night VLA~\citep{safenightvla2026} combine learned policies with
faster reactive modules, impedance adaptation, tactile calibration, or
safety filters during rollout. Their primary role is to stabilize or
constrain a learned policy at deployment time. Here, the controller
plays a different role. During collection, a deterministic
tactile reflex acts as a data-shaping teacher that changes which
contact trajectories are realized and recorded. At deployment,
this same reactive arbitration reappears as a decoupled gripper safety layer at runtime,
revealing what the tactile-free student did not inherit: high-rate
tactile correction under disturbance.

\section{Method}
\label{sec:method}

This section first formalizes in Fig.~\ref{fig:architecture} the
controller mechanism previewed in Fig.~\ref{fig:teaser} and specifies
how deployment is decoupled from collection
(Sec.~\ref{sec:method:overview}), then introduces the
controller-inspired dynamics-loss refinement used in Experiment~2
(Sec.~\ref{sec:method:secondary_refinement}), and finally summarizes
the evaluation quantities and tactile-regime summaries used throughout
the experiments (Sec.~\ref{sec:method:metrics}).

\subsection{Controller-Shaped Collection and Decoupled Deployment}
\label{sec:method:overview}

\begin{figure}[h]
  \centering
  \begin{tikzpicture}
    \node[
      draw=black!28,
      line width=0.35pt,
      inner sep=1.2pt,
      fill=white,
      drop shadow={shadow xshift=1.2pt, shadow yshift=-1.2pt,
        opacity=0.18}
    ] {\includegraphics[width=0.985\linewidth]{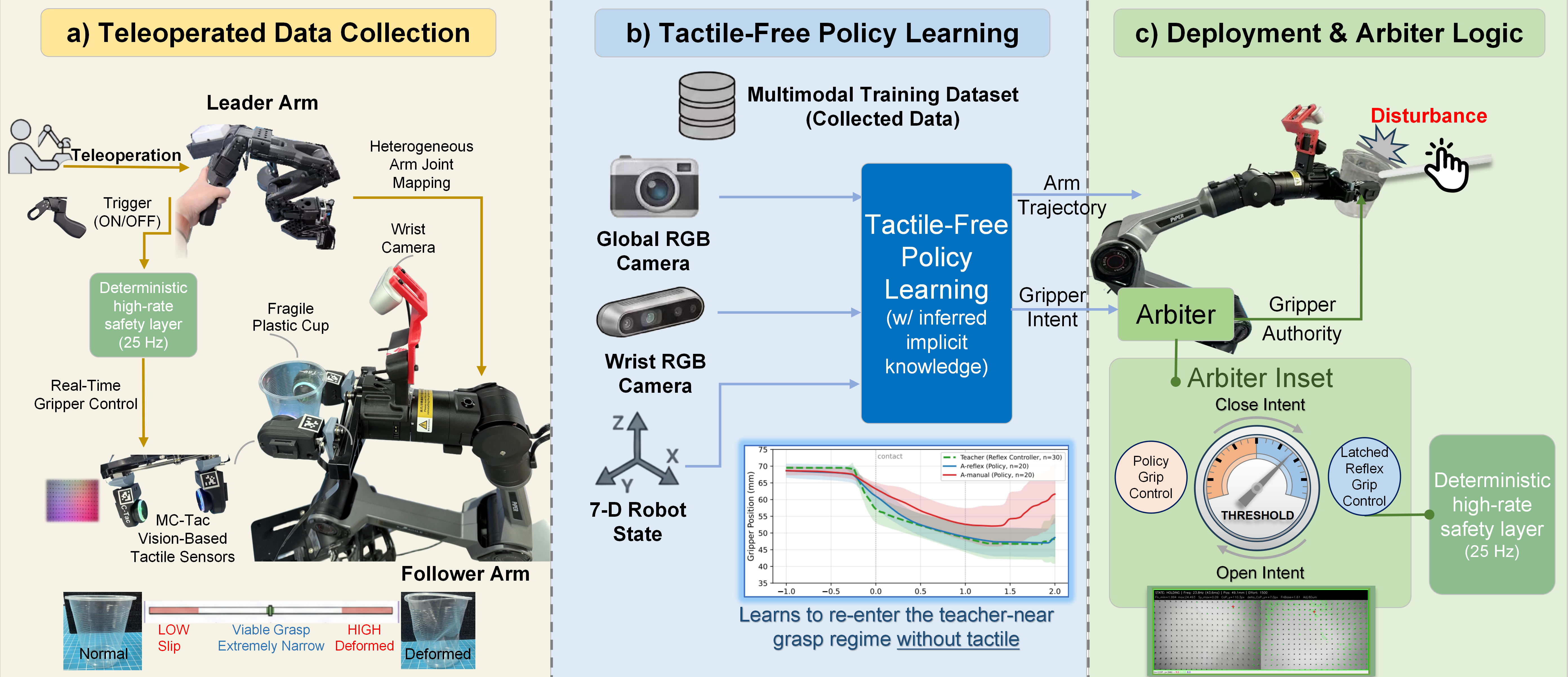}};
  \end{tikzpicture}
  \caption{\textbf{The controller matters twice}: as demonstration shaping during
  collection and as safety layer during deployment. During
  teleoperated collection, the reflex shapes demonstrations into a
  repeatable viable grasp regime; a tactile-free student then learns
  from RGB observations and robot state to re-enter that
  teacher-near regime without tactile input at inference; at
  deployment, arm control remains with the policy while a
  deterministic arbiter decouples gripper control by reinterpreting
  policy output as close/open intent, allowing the reflex to retain
  high-rate gripper authority for grasp stabilization, retention, and
  release during execution.}
  \label{fig:architecture}
\end{figure}

Fig.~\ref{fig:architecture} summarizes the paper's main mechanism.
The learned tactile-free student predicts arm trajectories and gripper
commands at chunked rates, whereas the deterministic tactile reflex operates
\emph{outside} the policy loop and regulates only the gripper at 25\,Hz
from dual tactile sensors. In this setup, the same controller appears in
two distinct roles: during collection it shapes the demonstrations seen
by behavioral cloning, and at deployment it can return as a local
high-rate gripper correction module.

\textbf{Collection-time role.}\label{sec:method:data}\quad During data
collection, one expert operator teleoperates a Piper
6-degree-of-freedom (6-DoF) arm via an SO101 leader arm. In both
collection modes, the operator has on-site visual access and a GUI
displaying live tactile parameters and slip events. In manual mode,
these signals are advisory only: the operator retains gripper authority
through coarse buttons, without haptic or controller actuation. In
reflex mode, TactileReflex~\citep{tactilereflex2025} autonomously
manages the follower gripper at 25\,Hz; the leader gripper acts only as
a binary intent signal, with closing arming the reflex and opening
releasing it. Teleoperation logging and dataset organization follow a
LeRobot-compatible format~\citep{lerobot2024}. For the reflex data,
each frame records robot state, tactile metrics, and camera streams,
but the 7-dimensional \emph{action} for imitation stores the follower's
\emph{realized} joint and gripper state rather than the leader's raw
command. The reflex dataset therefore records controller-shaped
realized trajectories. The original visually screened manual pool did
not retain tactile traces; a fresh contact-screened ablation retained
them only for offline ranking, never for actuation, ACT training, or
  inference. Collection-time gripper regulation is the primary
  data-source intervention, while object family, recorded action
  interface, and sensing stack remain fixed. Appendix~\ref{app:exp_controls}
  details manual-baseline screening, logging, and collection-order controls.

\textbf{Deployment-time role and decoupling.}\label{sec:method:reflex}\quad
The same TactileReflex controller can also remain active at deployment
as a deterministic 25\,Hz gripper controller~\citep{tactilereflex2025},
operating on normal force, shear slip,
and center of pressure from dual MC-Tac sensors~\citep{ren2023mctac}.
At deployment we use two runtime modes. In \emph{policy-only}, the
student controls both arm and gripper. In \emph{policy + arbiter}, the
same learned policy still controls the arm, but the gripper output
is reinterpreted as intent: a close command activates the reflex, which
then retains gripper authority despite policy jitter; release requires
sustained open intent plus a short suppression to avoid immediate
re-latching.
This decoupling matters because learned policies do not provide
high-rate tactile closed-loop correction inside the committed action
segment.\footnote{RDP~\citep{rdp2025} and
ImplicitRDP~\citep{implicitrdp2025} introduce a ``fast layer'' but keep
it learned and statistical.} In the current deployment setting, both
ACT and $\pi_{0.5}$ operate at much lower effective update rates than
the reflex and commit actions over finite chunks. The reflex instead
runs independently at about 25\,Hz, so it can intervene during grasp
stabilization and between policy updates. Appendix~\ref{app:arbiter}
gives the switch logic, Appendix~\ref{app:tactilereflex_engineering}
summarizes the engineering refinements used to sustain the controller
in this paper, and Appendix~\ref{app:exp_setup} reports the
implementation details and rollout frequencies.
\subsection{Dynamics-Loss Refinement for Controller-Shaped Regime Entry and Release}
\label{sec:method:secondary_refinement}

After establishing the collection-time mechanism, Experiment~2 tests a
controller-inspired dynamics-loss refinement on reflex-data training.
The goal is specific: improve how decisively the learned policy enters
and exits the controller-shaped nominal contact regime.

Concretely, we add a gripper dynamics consistency term during training
as a simple second-order trajectory-shape consistency prior on the demonstrated
gripper trajectory. This is analogous to smoothness costs used in robot
trajectory optimization~\citep{ratliff2009chomp}, but here it is used
only as a narrow train-time prior on gripper trajectory shape.
Let $g_{1:T}$ denote the demonstrated gripper trajectory inside an
action chunk and $\hat{g}_{1:T}$ the policy prediction. In addition to
the standard training objective $\mathcal{L}_{\mathrm{main}}$, we
penalize mismatch in the second-order temporal difference of the
gripper trajectory,
\[
\mathcal{L}_{\mathrm{dyn}} =
\frac{1}{T-2}\sum_{t=2}^{T-1}
\left|\Delta^2 \hat{g}_t - \Delta^2 g_t\right|,
\qquad
\Delta^2 g_t = g_{t+1} - 2g_t + g_{t-1},
\]
and optimize
\[
\mathcal{L}_{\mathrm{total}} =
\mathcal{L}_{\mathrm{main}} + \lambda_{\mathrm{dyn}}\,
\mathcal{L}_{\mathrm{dyn}}.
\]
In all Experiment~2 runs, we use $\lambda_{\mathrm{dyn}} = 0.1$.
We use the second-order difference because it captures how the gripper
trajectory bends as it approaches, enters, and settles into contact,
rather than whether action values are close at each step. The term
therefore biases the policy toward the teacher's contact-to-hold
trajectory shape and transition sharpness, not just pointwise gripper
positions. This refinement is training-only: it adds no tactile input
at inference and does not alter the rollout or arbiter logic.

\subsection{Episode Outcomes and Tactile-Regime Evaluation Metrics}
\label{sec:method:metrics}

We evaluate both whole-episode task outcomes and a
teacher-referenced tactile-regime diagnostic. The episode-level
outcomes are $\texttt{stable}$, $\texttt{loose}$, and
$\texttt{drop}$, which summarize visual grasp quality over the full
episode. The tactile diagnostic is defined from a
contact-qualified (CQ) predicate and a teacher-near band around the reflex
stable-hold reference. In the main text, we report three summaries:
\emph{CQ episodes} as a conservative regime-entry check, the
event-style \emph{$\geq$10-frame teacher-near run} as the primary
summary of regime entry and sustained re-entry, and
\emph{CQ frames near teacher} as a secondary occupancy-based summary.
Because \emph{CQ episodes} is derived from a conservative frame-level
tactile contact criterion, tactile regime entry need not coincide with
visually stable completion.

Experiment~2 additionally reports
\emph{CQ-onset 3\,s near/CQ occupancy} for early teacher-near entry
after meaningful contact begins, together with \emph{time to release}
and \emph{release-time reduction}. Full numerical thresholds and
formal definitions are deferred to
Appendix~\ref{app:contact_threshold} and
Appendix~\ref{app:exp_metrics}, and matched-comparison controls are
given in Appendix~\ref{app:exp_controls}.

\section{Experiments}
\label{sec:experiments}

This section evaluates whether controller-shaped collection improves
what a tactile-free student learns under ID conditions and explores
whether the same data-source effect extends to an unseen paper-cup OOD
variant (Sec.~\ref{sec:exp:data_quality}); the latter is treated as
exploratory scope evidence because its small-sample difference is not
statistically significant. A small
train-time refinement sharpens entry and release within the learned
nominal regime (Sec.~\ref{sec:exp:refinement}); and randomized
disturbance exposes the residual need for deployment-time arbitration
(Sec.~\ref{sec:exp:hierarchical}).

\subsection{Setup}
\label{sec:exp:setup}

\textbf{Platform and object.}\quad
We use a Piper 6-degree-of-freedom (6-DoF) arm with a 1-DoF gripper,
so the recorded robot state is 7-DoF, together with dual MC-Tac
vision-based tactile sensors~\citep{ren2023mctac}, wrist and front RGB
cameras, and a disposable plastic cup (3.5\,g, 0.3\,mm wall,
narrow sub-Newton deformation regime). Experiment~1a uses
ACT~\citep{zhao2023act} as the primary controlled testbed;
Experiments~1b, 1c, and~3 use the tactile-free $\pi_{0.5}$
model~\citep{pi05_2025}; and Experiment~2 reports matched refinement
comparisons on both. Detailed hardware and model specifications are
deferred to Appendix~\ref{app:exp_setup}.

\textbf{Evaluation conditions.}\quad
We evaluate nominal-workspace ID grasping with
randomized cup placement, an unseen paper-cup OOD setting, and a
randomized external-disturbance test during place. Episode labels,
tactile-regime metrics, and paired-comparison invariants follow
Sec.~\ref{sec:method:metrics}. Unless otherwise stated, main-text
tables report one representative run; additional-seed
robustness checks are summarized in the result paragraphs and reported
in Appendix~\ref{app:multiseed}. Appendix~\ref{app:exp_design}
collects matched-comparison controls, metric roles, failure taxonomy,
and statistical notes.

\subsection{Experiment 1: Learning from Controller-Shaped Demonstrations}
\label{sec:exp:data_quality}

Experiment~1 tests the main learning claim: if controller-shaped demonstrations 
improve what a tactile-free student can learn, then a student trained on reflex-data 
demonstrations should outperform a matched student trained on visually screened manual
demonstrations under the same deployment conditions.
Experiment~1a establishes this controlled comparison on ACT as the
primary test, Experiment~1b checks whether the same
data-source effect carries to a different backbone, and
Experiment~1c probes the same data-source effect on an unseen
paper-cup OOD variant.

\subsubsection{Experiment 1a: ACT as the Primary Controlled Test}

\textbf{Protocol.}\quad
Two tactile-free ACT instances differ only in training data source: the
\emph{reflex-data policy} uses 30 episodes collected with the reflex
active at 25\,Hz, whereas the \emph{visually screened manual-data policy} uses
30 demonstrations selected from 100 manual trials under the same
visually guided teleoperation protocol and stable-outcome screening
rule. Object, cameras, training budget, architecture, training recipe,
and evaluation protocol are otherwise matched, so the comparison
isolates collection-time gripper regulation. Both are evaluated under
nominal ID conditions ($20$ trials each);
Appendix~\ref{app:exp_controls} details the manual-pool construction.

\begin{table}[h]
  \centering
  \caption{ACT ID deployment ($20$ trials per row). Rows~1--2 compare
  training data under policy-only runtime; row~3 adds the arbiter to
  the same visually screened manual-data policy.}
  \label{tab:exp1}
  \footnotesize
  \begin{tabular}{llccc}
    \toprule
    \textbf{Training data} & \textbf{Runtime mode} & \textbf{Stable}
      & \textbf{Loose} & \textbf{Drop} \\
    \midrule
    Reflex-data & policy-only & \textbf{95\% (19/20)} & 5\% (1/20) & \textbf{0\% (0/20)} \\
    Visually screened manual-data & policy-only & 15\% (3/20) & 55\% (11/20) & 30\% (6/20) \\
    Visually screened manual-data & policy + arbiter & \textbf{50\% (10/20)} & 20\% (4/20) & 30\% (6/20) \\
    \bottomrule
  \end{tabular}
\end{table}

\textbf{Main ACT result.}\quad
Table~\ref{tab:exp1} shows a sharp training-data contrast: the
reflex-data ACT policy achieves 95\% (19/20) stable grasps with zero
drops, whereas the visually screened manual-data policy reaches only 15\%
(3/20) stable with a 30\% (6/20) drop rate
($p = 4.1 \times 10^{-7}$, two-sided Fisher exact test). Adding the
runtime arbiter to the same manual-data policy raises stability to
50\% (10/20) but leaves drops unchanged at 30\% (6/20), so runtime
arbitration rescues some marginal grasps without recovering the
nominal quality supplied by controller-shaped collection. Additional
seeds preserve the same ranking (95.0$\pm$4.1\%,
16.3$\pm$4.8\%, and 48.8$\pm$2.5\% stable;
Appendix~\ref{app:multiseed}).

\textbf{Contact-screened controlled ablation.}\quad
As a separate single-run check, we collected 100 fresh manual
demonstrations with tactile logging and selected the top 30 by the same
teacher-referenced CQ-near criterion used for analysis. The logged
tactile signals were used only for offline ranking: they neither
actuated the manual gripper nor entered ACT training or inference. The
resulting contact-screened manual-data ACT policy achieved 30\% (6/20)
stable outcomes versus 95\% (19/20) for the reflex-data policy
($p=3.9\times10^{-5}$, two-sided Fisher exact test). This controlled
ablation supports the collection-time advantage after contact-quality
screening, but, because it uses one training and evaluation run, is not
a multi-seed robustness claim. See Appendix~\ref{app:contact_screened}
for full details.

\begin{figure}[h]
  \centering
  \includegraphics[width=0.95\linewidth]{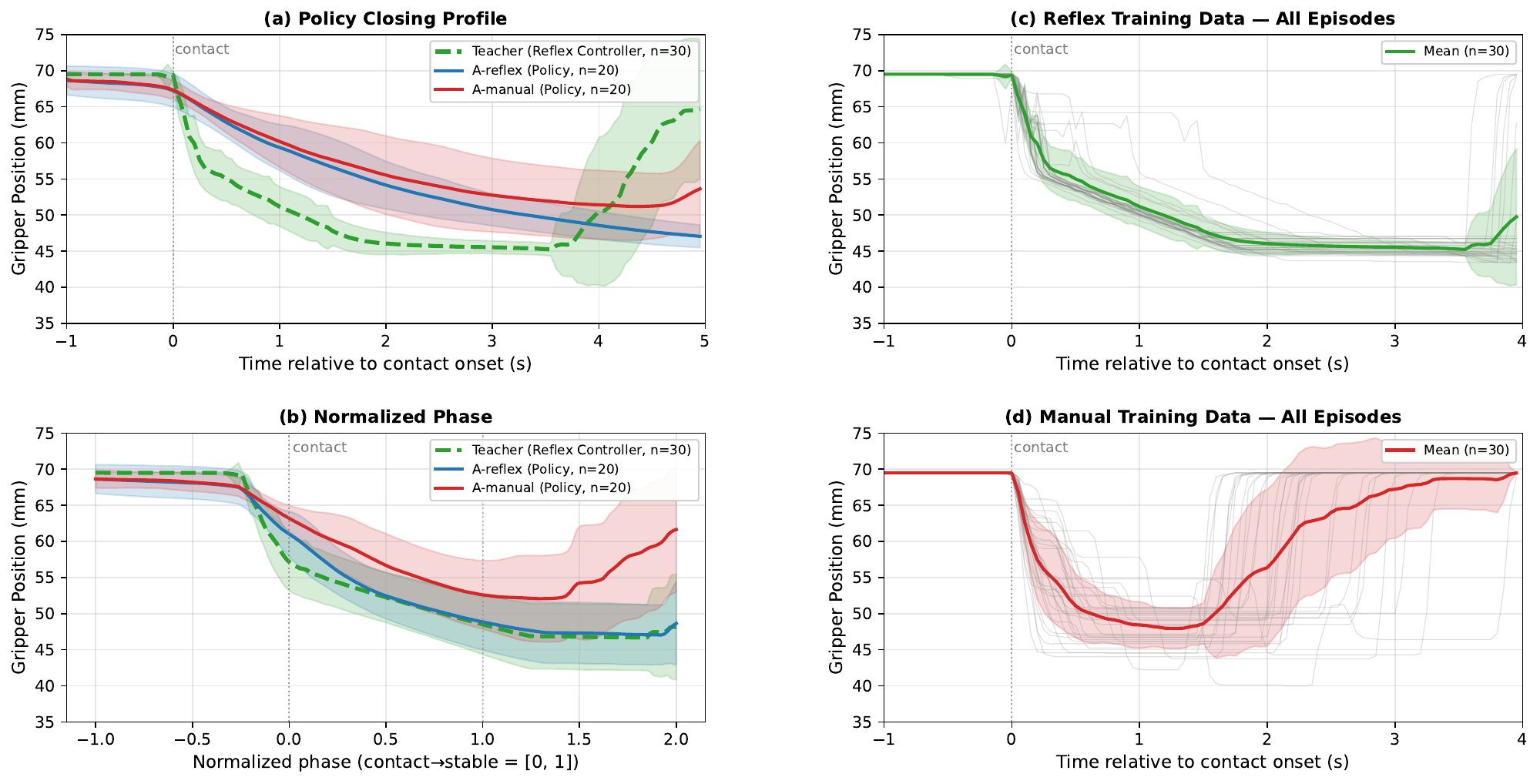}
  \caption{Controlled comparison on ACT. The reflex-data
  ACT policy tracks the teacher through contact-to-hold, whereas the
  visually screened manual-data ACT policy deviates after closure and misses the
  teacher-defined nominal regime. Some visually screened manual demonstrations
  enter the viable closing range during collection, but the learned
  manual-data policy does not reliably re-enter that regime at
  deployment. Grey: individual rollouts; color: mean $\pm 1\sigma$.}
  \label{fig:killer_plot}
\end{figure}

\begin{table}[h]
  \centering
  \caption{Tactile-regime diagnostic for ACT under ID deployment ($20$
  trials each). Metrics follow Sec.~\ref{sec:method:metrics}. Because
  \emph{CQ frames near teacher} is computed only on CQ frames, high
  values in rows with very few CQ episodes do not imply regime parity
  (Appendix~\ref{app:exp_sparsecq}).}
  \label{tab:exp1a_act_tactile}
  \footnotesize
  \setlength{\tabcolsep}{2pt}
  \resizebox{\linewidth}{!}{%
  \begin{tabular}{lccc}
    \toprule
    \textbf{ACT policy} & \textbf{CQ episodes}
      & \textbf{$\geq$10f teacher-near run}
      & \textbf{CQ frames near teacher} \\
    \midrule
    Reflex-data policy & \textbf{19/20} & \textbf{15/20} & \textbf{68.7\%} \\
    Visually screened manual-data policy & 1/20 & 0/20 & 12.5\% \\
    Visually screened manual-data policy + arbiter & \textbf{10/20} & \textbf{7/20} & \textbf{39.4\%} \\
    \bottomrule
  \end{tabular}%
  }
\end{table}

\textbf{Trajectory diagnostic.}\quad
Fig.~\ref{fig:killer_plot} and
Table~\ref{tab:exp1a_act_tactile} show the mechanism behind this gap: controller-shaped 
demonstrations let the reflex-data ACT policy repeatedly recover the teacher-defined 
nominal contact regime, whereas the visually screened manual-data policy rarely does;
the arbiter yields only partial recovery. Controller-shaped collection 
therefore makes the narrow viable regime learnable, whereas careful manual
collection without closed-loop tactile actuation does not reliably encode the
same controller-regulated behavior across contact, lift, transport, and release.

\subsubsection{Experiment 1b: $\pi_{0.5}$ Cross-Backbone ID Validation}

\textbf{ID corroboration on $\pi_{0.5}$.}\quad
On $\pi_{0.5}$, the same separation persists under ID
conditions: the reflex-data policy reaches 95\% (19/20) stable
grasps, whereas the visually screened manual-data policy reaches 5\% (1/20)
stable and the same policy with the arbiter recovers only to 55\%
(11/20) (Table~\ref{tab:exp1b_pi}). Appendix~\ref{app:exp_pi05_tactile}
reports the matching teacher-referenced tactile separation, again with
only partial arbiter recovery. Additional seeds preserve the same
ranking (96.3$\pm$4.8\%, 6.3$\pm$4.8\%, and 52.5$\pm$6.5\% stable;
Appendix~\ref{app:multiseed}); exact tests are in
Appendix~\ref{app:exp_stats}.

\begin{table}[h]
  \centering
  \caption{$\pi_{0.5}$ task outcomes under ID and unseen paper-cup OOD
  conditions. The upper block reports ID results (Exp.~1b); the lower
  block reports unseen paper-cup OOD results (Exp.~1c).}
  \label{tab:exp1b_pi}
  \footnotesize
  \begin{tabular}{lccc}
    \toprule
    \rowcolor{black!8}
    \textbf{$\pi_{0.5}$ condition / variant} & \textbf{Stable}
      & \textbf{Loose} & \textbf{Drop} \\
    \midrule
    \rowcolor{black!4}
    \multicolumn{4}{l}{\textbf{ID}} \\
    ID, reflex-data policy & \textbf{95\% (19/20)} & 5\% (1/20) & \textbf{0\% (0/20)} \\
    ID, visually screened manual-data policy & 5\% (1/20) & 70\% (14/20) & 25\% (5/20) \\
    ID, visually screened manual-data policy + arbiter & \textbf{55\% (11/20)} & 25\% (5/20) & 20\% (4/20) \\
    \rowcolor{black!4}
    \multicolumn{4}{l}{\textbf{Paper cup}} \\
    Paper cup, reflex-data policy & \textbf{80\% (8/10)} & 20\% (2/10) & \textbf{0\% (0/10)} \\
    Paper cup, visually screened manual-data policy & 30\% (3/10) & 50\% (5/10) & 20\% (2/10) \\
    \bottomrule
  \end{tabular}
\end{table}

\subsubsection{Experiment 1c: Unseen Paper-Cup OOD Extension}
\label{sec:exp:modality}

\textbf{OOD extension on the unseen paper cup.}\quad
Using the paper-cup block of Table~\ref{tab:exp1b_pi}, the same
mechanism is tested on an unseen paper cup. Because this cup is
mechanically and visually more forgiving than the training cup, the
result is treated as exploratory OOD scope evidence rather than as a second
confirmatory claim. The reflex-data policy is numerically higher
(80\% stable vs.\ 30\%), but the difference is not statistically
significant (two-sided Fisher $p=0.0698$). Appendix~\ref{app:exp_papercup} records the
setup and exact-test note, and Appendix~\ref{app:act_tactile} records
the preliminary tactile-image checks that motivate centering the main
study on a tactile-free student.

\subsection{Experiment 2: Controller-shaped behavior refinement with Dynamics-Loss}
\label{sec:exp:refinement}

Experiment~2 is a secondary follow-up to Experiment~1: with backbone,
sensing, runtime, and deployment interface fixed, can a small
train-time refinement sharpen entry and release behavior within the learned
nominal regime?

\begin{table}[h]
  \centering
  \caption{Reflex-data policy comparison with and without the dynamics consistency loss.}
  \label{tab:exp2_dyn}
  \scriptsize
  \setlength{\tabcolsep}{2pt}
  \renewcommand{\arraystretch}{1.0}
  \begin{tabular}{llcccc}
    \toprule
    \textbf{Backbone} & \textbf{Variant}
      & \shortstack{\textbf{$\geq$10f} \textbf{run} }
      & \shortstack{\textbf{CQ-onset 3\,s} \textbf{near/CQ} }
      & \shortstack{\textbf{Time to} \textbf{release} }
      & \shortstack{\textbf{Release-time} \textbf{reduction} } \\
    \midrule
    ACT & baseline & 15/20 & 72.2\% & 15.7\,s & --- \\
    ACT & + dyn.\ loss & \textbf{18/20} & \textbf{82.7\%} & \textbf{13.4\,s} & \textbf{14.6\%} \\
    $\pi_{0.5}$ & baseline & 12/20 & 78.5\% & 28.1\,s & --- \\
    $\pi_{0.5}$ & + dyn.\ loss & \textbf{16/20} & \textbf{85.3\%} & \textbf{21.7\,s} & \textbf{22.8\%} \\
    \bottomrule
  \end{tabular}
\end{table}

\textbf{Result and interpretation.}\quad
Table~\ref{tab:exp2_dyn} shows the same refinement trend on both
backbones: teacher-near runs rise from 15/20 to 18/20 on ACT and from
12/20 to 16/20 on $\pi_{0.5}$, with higher onset occupancy and earlier
release in both cases. We interpret this narrowly as sharper regime
entry and earlier release onset within the same deployment interface.
Additional seeds preserve this CQ-onset improvement on both
backbones (ACT: 72.3$\pm$1.6\% to 82.6$\pm$1.4\%;
$\pi_{0.5}$: 78.4$\pm$0.9\% to 85.6$\pm$1.5\%;
Appendix~\ref{app:multiseed}); Appendix~\ref{app:dyn_refinement}
discusses the omitted full-horizon \emph{teacher-nearness} diagnostic.

\subsection{Experiment 3: Exploring the Boundary of Tactile-Free Policy}
\label{sec:exp:hierarchical}

Experiment~3 tests the deployment-time boundary of tactile-free policy:
under external disturbance, a tactile-free policy with non-negligible
inference-to-action delay may still require a decoupled real-time gripper controller,
even when trained on reflex-data.

\textbf{Protocol.}\quad
Once the cup is stably lifted and place begins, a second person
applies a lateral stick disturbance over 20 randomized trials per
mode. In policy + arbiter mode, the arm remains policy-controlled
while gripper authority is latched to the reflex after grasp intent
and released only after sustained open intent
(Fig.~\ref{fig:exp3_disturbance}), preserving the intended
arm--gripper decoupling during disturbance.

\begin{wraptable}{r}{0.49\textwidth}
  \vspace{-0.9\baselineskip}
  \centering
  \scriptsize
  \setlength{\tabcolsep}{2pt}
  \begin{tabular}{lcc}
    \toprule
    \textbf{Runtime mode}
      & \textbf{Retained} & \textbf{Failure} \\
    \midrule
    Policy-only & 55\% (11/20) & 45\% (9/20) \\
    Policy + arbiter & \textbf{100\% (20/20)} & \textbf{0\% (0/20)} \\
    \bottomrule
  \end{tabular}
  \caption{Runtime comparison on the same reflex-data
  $\pi_{0.5}$ policy ($20$ disturbed trials per mode).}
  \label{tab:exp3_boundary}
  \vspace{-0.2\baselineskip}
\end{wraptable}

\textbf{Result and interpretation.}\quad
Table~\ref{tab:exp3_boundary} shows that, on the same policy, policy-only control fails in 9/20 disturbance
trials, whereas deterministic reflex arbitration retains all 20/20
($p = 1.23 \times 10^{-3}$, two-sided Fisher exact test). Because the
learned policy is identical in both modes, the gap is due to
runtime control rather than planning: the tactile-free student
transfers nominal task behavior, but high-rate correction inside a
committed chunk remains controller-dependent under disturbance.
Additional seeds show
the same split (52.5$\pm$6.5\% vs.\ 98.8$\pm$2.5\% retention;
Appendix~\ref{app:multiseed}).

\section{Conclusion}
\label{sec:conclusion}

We show that a deterministic tactile reflex controller can play a distinct
collection-time role by producing demonstrations with controller-shaped
grasping behavior, providing a more consistent training distribution for
tactile-free policy learning. This effect is clearest on ACT, carries over to
$\pi_{0.5}$ under ID conditions, and shows a favorable exploratory trend on
an unseen paper-cup OOD setting. At deployment, high-rate tactile arbitration
plays a complementary role: it provides rapid grasp correction under external
disturbance that is unavailable from policy-only execution. Together, these
results position tactile feedback not only as a policy input, but also as a
teacher for shaping demonstrations and as a fast reactive layer that
complements learned manipulation policies.

\section{Limitations and Future Work}
\label{sec:limitations}

The evidence in this paper is narrow. We evaluate only disposable
cups, one unseen paper-cup OOD setting, and a single disturbance
family, so the claim is strongest for direct-contact fragile
containers whose success depends on maintaining a narrow gripper-force
window. The contact-screened controlled ablation contains only one ACT
training and evaluation run and no additional-seed study, so it is
corroborative rather than a robustness result. Both collection modes
gave the operator visual access to the GUI tactile cues, but manual
control still incurred observation, reaction, and button-execution
delay. Moreover, one expert operator collected the data, reflex data
preceded the manual pools, and collection order was not randomized.
Practice would tend to favor the later manual data, but operator and
temporal effects remain possible. We do not test broader contact-rich tasks where arm motion,
object geometry, or tool interaction matter more, and Experiment~3
should be read as a reactive-boundary demonstration rather than a
calibrated robustness curve. Our findings are similarly specific:
controller-shaped demonstrations make a viable nominal contact regime
more collectable and learnable, but they do not close the deployment-time
control gap once contact leaves that regime. The main limitation is
insufficient high-rate recovery under disturbance: the deployed
explicit tactile real-time controller can preserve gripper-side
behavior, but it cannot provide closed-loop real-time coordination of
the whole arm. A next step is to scale this paradigm to more object
families and disturbances, and to pair it with richer arm--gripper
visuo-tactile coordination.


\clearpage
\acknowledgments{The authors thank the Area Chair and the anonymous reviewers for their constructive feedback. This work is partially supported by the Research Travel Grant of the Base of Red Bird MPhil (RBM) at The Hong Kong University of Science and Technology (Guangzhou). We would like to thank our RBM Project Supervisor Jinni Zhou for the academic support.}


\bibliography{references}

\clearpage
\appendix
\section{Appendix-Only Notes on Naive Tactile-Image Fusion}
\label{app:act_tactile}

This appendix documents appendix-only naive tactile-image fusion
variants referenced in Sec.~\ref{sec:exp:modality}. These variants test a naive
direct visuo-tactile image-fusion setup and are included only to show
what this formulation did and did not provide under the current data
and sensing setup. They are outside the main system and are not a
controlled comparison with richer tactile-fusion methods.

\paragraph{Motivation for the collection-time formulation.}
These preliminary variants motivated the paper's focus. Direct
tactile-image fusion did not yield a reliable improvement in our setup,
whereas policies trained on reflex-shaped demonstrations retained strong
nominal task performance without tactile input at inference. This
observation led us to study tactile sensing as a collection-time
mechanism for shaping demonstrations, rather than as an additional
policy input.

\paragraph{ACT tactile-image extension.}
An ACT variant was evaluated that augments the vision-only input with
left and right tactile image streams as additional ResNet-18 image
inputs. Under the current setup, this naive direct-fusion formulation
did not effectively improve policy learning. Instead, it caused severe
degradation: the policy often lost usable task behavior and could fail
to maintain stable pick-and-place performance.

\paragraph{$\pi_{0.5}$ tactile-image variant.}
A tactile-image variant of $\pi_{0.5}$ was also evaluated by adding the
left and right tactile views as extra image inputs under the same
rollout setting used for the main $\pi_{0.5}$ study. Unlike ACT,
$\pi_{0.5}$ retained basic task ability and could still complete the
task, but it did not show clear or reliable improvement over the
vision-only baseline in success rate or grasp quality.

\paragraph{Interpretation.}
Taken together, these diagnostics suggest that directly feeding
visuo-tactile image streams into policy training did not effectively
improve fine-tuning in the current setup. One plausible explanation is
that raw tactile-image streams are noisy and difficult to exploit
through this straightforward extra-camera integration, but no
definitive mechanism is claimed here. More structured visuo-tactile
learning may still be promising and requires further study. The
negative effect was stronger for ACT, while $\pi_{0.5}$ remained
functional but did not show clear gains. We treat these results as
appendix-only boundary evidence for this naive direct-fusion
formulation, not as a statement about tactile information in general or
about richer tactile integration schemes. They justify the deployment
choice studied in this paper, but they are not themselves a main
experimental claim.

\clearpage
\section{Deployment-Time Arbiter Logic}
\label{app:arbiter}

Sec.~\ref{sec:method:reflex} gives the main-paper description of the
deployment-time arbiter. This appendix records the switch logic more
explicitly. Here, $g_t$ denotes the policy gripper intent, $r_t$
indicates whether the reflex is latched, $s_t$ denotes the temporary
release-suppression state, and $K$ is the number of consecutive
open-intent policy ticks required for release. The symbols
$g_{\mathrm{close}}$ and $g_{\mathrm{open}}$ are the close and open
thresholds, respectively. The arbiter changes only gripper authority;
the arm remains policy-controlled throughout. Algorithm~\ref{alg:arbiter}
summarizes the latching, release, and suppression logic used in the
policy+arbiter runtime mode.

\par\smallskip
\noindent
\refstepcounter{algobox}\label{alg:arbiter}
\fbox{%
  \begin{minipage}{0.96\linewidth}
    \footnotesize
    \textbf{Algorithm \thealgobox. Deployment-time gripper arbitration}

    \vspace{2pt}
    \begin{algoic}
      \Require{Runtime mode $m \in \{\texttt{policy-only},\, \texttt{policy+arbiter}\}$;
      policy arm action $a_t^{\mathrm{arm}}$; policy gripper output
      $g_t$; reflex-active flag $r_t \in \{0,1\}$;
      release-suppression flag $s_t \in \{0,1\}$; close threshold
      $g_{\mathrm{close}}$; open threshold $g_{\mathrm{open}}$;
      open-intent horizon $K$}
      \State{Execute $a_t^{\mathrm{arm}}$ from the policy.}
      \If{$m = \texttt{policy-only}$}
        \State{Apply $g_t$ directly to the gripper and return.}
      \ElsIf{$r_t = 1 \land g_t \geq g_{\mathrm{open}}$ for $K$
        consecutive policy ticks}
        \State{Send \texttt{stop} to the reflex and return gripper
        authority to the policy.}
      \ElsIf{$r_t = 1$}
        \State{Execute the reflex gripper command at 25\,Hz and
        return.}
      \ElsIf{$s_t = 1 \land g_t \geq g_{\mathrm{open}}$}
        \State{Apply $g_t$ directly and suppress immediate re-latching.}
      \Else
        \If{$g_t \leq g_{\mathrm{close}}$}
          \State{Send \texttt{start} to the reflex.}
        \EndIf
        \If{the reflex has latched after the start request}
          \State{Execute the reflex gripper command.}
        \Else
          \State{Apply $g_t$ directly to the gripper for this policy
          step.}
        \EndIf
      \EndIf
    \end{algoic}
  \end{minipage}%
}
\par\smallskip

\section{Implementation Notes for the Deployed Reflex Controller}
\label{app:tactilereflex_engineering}

This appendix summarizes implementation details of the deployed reflex
controller used in this paper. These details are engineering-oriented
and do not change the controller logic described
in~\citep{tactilereflex2025}. The tactile metrics, state machine, and
gripper decision rules are
unchanged; only the runtime pipeline was modified to improve
operational stability at the target control rate.

\paragraph{What remained unchanged.}
The controller still uses the same three-channel tactile logic as the
base formulation: onset detection from the weaker-side normal-contact
signal, slip correction from shear cues, and over-compression relief
from center-of-pressure cues, with the normal-force protection branch
retaining highest priority. We therefore do not treat the controller
itself as a new algorithmic contribution.

\paragraph{Asynchronous tactile acquisition.}
Image capture is decoupled from the control loop through persistent
background acquisition threads. The controller consumes the latest
available preprocessed tactile frames rather than synchronously waiting
for a fresh read at each step. This reduces blocking camera latency and
improves loop regularity.

\paragraph{Parallel left--right metric computation.}
Left and right tactile metrics are computed in parallel rather than
serially. The underlying processing pipeline remains the same---frame
differencing against a baseline, contact-mask construction, weighted
normal-contact estimation, optical-flow-based shear estimation, and
center-of-pressure extraction---but the two sensors are processed
concurrently at each step. This improves wall-clock throughput without
changing the controller inputs.

\paragraph{Separated calibration and streaming.}
Baseline and noise calibration are performed in a dedicated
initialization phase before live streaming begins. The controller first
collects multiple non-contact frames per sensor, estimates a multi-frame
baseline image, and sets the per-sensor contact threshold from the
upper-percentile intensity-difference statistics. Live threaded
streaming starts only after calibration succeeds.

\paragraph{Runtime monitoring and logging.}
The implementation records per-step wall-clock timing, moving-average
loop frequency, controller state, tactile statistics, and discrete
corrective actions to comma-separated value (CSV) logs. These logs are not part of the control
law, but they support debugging and rate verification.

\paragraph{Role in this paper.}
We include these refinements for completeness because the paper relies
on a controller whose mechanism is fixed and operationally
repeatable. The point is not that these changes alter what
TactileReflex computes, but that they make the same controller more
practical to run consistently in the collection-time-teacher setting
studied here.

\section{Contact-Qualified Threshold for the Tactile Diagnostic}
\label{app:contact_threshold}

Sec.~\ref{sec:method:metrics} defines the tactile-regime diagnostic
used in Experiment~1. This appendix only explains the provenance of the
contact-qualified threshold. We denote by $F_{n,\min}$ the smaller of
the two normal-contact proxies from the left and right tactile sensors.
Here $F_n$ is a dimensionless vision-tactile proxy; it has not been
calibrated in Newtons.
We use the weaker side rather than $F_{n,\max}$ or an average because,
for fragile grasps, loss of meaningful contact is usually governed by
the bottleneck side. This makes $F_{n,\min}$ the more conservative
proxy for whether the grasp has actually entered a usable contact
state.

The threshold $F_{n,\min}>1.5$ is not meant to define the teacher-near
band itself, nor is it chosen by tuning on policy success. Its role is
only to exclude incidental light touch and retain frames that have
entered a minimally meaningful grasp-contact regime. In other words, it
defines the lower bound for a contact-qualified (CQ) frame, not the
teacher-near operating range.

This lower bound is tied to the TactileReflex controller scale used
in~\citep{tactilereflex2025} rather than introduced as a new free
hyperparameter. In that controller formulation, the transition from
grasp closing to holding was
triggered once the weaker-side normal-contact proxy reached a small
positive contact threshold on this same scale \citep{tactilereflex2025}.
In the controller configuration used throughout this paper, that
grasp-onset threshold is 1.5, so the same value is reused here as a
conservative CQ lower bound. This continuity gives the threshold a
controller-side physical interpretation: it marks ``contact has become
meaningful enough to count as grasp onset'' rather than ``the policy is
already in the teacher-near band.''

The teacher-near regime is defined separately. After applying the CQ
filter, $F_{n,\min}$ is compared to the reflex-cup stable-hold reference
statistics. For all teacher-near quantities reported in the main text
and appendix tables---including \emph{$\geq$10-frame teacher-near run},
\emph{CQ frames near teacher}, and \emph{CQ-onset 3\,s near/CQ
occupancy}---the same stable-hold teacher-near band is used
($2.41\pm0.7$, dimensionless, in the main-text visualization). A stricter teacher
$p10$--$p90$ range is kept only as an appendix-side alternative
sensitivity reference and is not used for the reported results. The
logic is intentionally two-stage, and the two stages serve different
purposes:
\emph{(1)}~$F_{n,\min}>1.5$ asks whether the policy has established
meaningful grasp contact at all; \emph{(2)}~the teacher-referenced band
asks whether that contact lies near the stable regime shaped by the
controller demonstrations.

Separate external-loading observations characterize the cup material,
not the tactile proxy. Visible deformation appeared near an applied
load of approximately $0.5\,\mathrm{N}$, and sustained loading
approaching but below $1\,\mathrm{N}$ produced irreversible deformation
that required manual reshaping. These external-load values describe the
material's response only; they do not calibrate $F_n$, the CQ threshold,
the teacher-near band, or any controller threshold.

\section{Additional Experimental Details}
\label{app:exp_details}

This appendix records secondary experimental details that support the
main paper but are not part of its primary evidence chain. In
particular, it makes the matched-comparison controls, metric roles,
failure taxonomy, and statistical notes explicit. For readability, it
groups these details into setup, comparison design, additional-seed
robustness, and supplementary Experiment~1 notes.

\subsection{Platform, Models, and Policy Variants}
\label{app:exp_setup}

\paragraph{Platform and model details.}
The platform uses a Piper 6-degree-of-freedom (6-DoF) arm, dual MC-Tac
vision-based tactile sensors, an Intel RealSense wrist camera, and an
icspring front camera; all training and inference run on a single
RTX~5090. Teleoperation logging and dataset storage follow a
LeRobot-compatible format~\citep{lerobot2024}. Action Chunking with
Transformers (ACT) uses an 18M policy with a ResNet-18 visual encoder
and, in the current rollout setting, a 15\,Hz full policy-update loop.
The main tactile-free $\pi_{0.5}$ policies use front and wrist
red-green-blue (RGB) camera streams plus a 7-D robot state and are
built on a 2B+ vision-language-action (VLA) model with SigLIP vision
transformer (ViT) and Gemma, fine-tuned with Low-Rank Adaptation (LoRA)
at 10\,Hz. In the main deployment study, the deployed $\pi_{0.5}$
policy executes action steps at 20\,Hz under a 10-step full-chunk
synchronous rollout, which corresponds to an empirical wall-clock
policy re-query rate of about 1.7--1.8\,Hz in the recorded rollouts.
The tactile reflex operates independently at 25\,Hz.

\paragraph{Policy variants.}
Table~\ref{tab:configs} summarizes the policy variants referenced in
the main paper.

\begin{table}[h]
  \centering
  \caption{Policy variants. Experiment~1 changes training data,
  Experiment~2 the training objective, and Experiment~3 runtime control
  under randomized external disturbance; the final block lists
  appendix-only tactile-image variants. All use the same 7-D robot
  state and front/wrist RGB views.}
  \label{tab:configs}
  \scriptsize
  \setlength{\tabcolsep}{3pt}
  \renewcommand{\arraystretch}{1.03}
  \begin{tabular}{llll}
    \toprule
    \textbf{Model} & \textbf{Variant} & \textbf{Tactile input} & \textbf{Gripper control} \\
    \midrule
    \multicolumn{4}{l}{\emph{Experiment 1 (controller-shaped demonstrations):}} \\
    ACT (18M) & reflex-data policy & none & policy-only \\
    ACT (18M) & visually screened manual-data policy & none & policy-only \\
    ACT (18M) & visually screened manual-data policy + arbiter & none & policy + arbiter \\
    $\pi_{0.5}$ (2B+) & reflex-data policy & none & policy-only \\
    $\pi_{0.5}$ (2B+) & visually screened manual-data policy & none & policy-only \\
    $\pi_{0.5}$ (2B+) & visually screened manual-data policy + arbiter & none & policy + arbiter \\
    \midrule
    \multicolumn{4}{l}{\emph{Experiment 2 (dynamics-loss refinement):}} \\
    ACT (18M) & reflex-data + dyn. loss & none & policy-only \\
    $\pi_{0.5}$ (2B+) & reflex-data + dyn. loss & none & policy-only \\
    \midrule
    \multicolumn{4}{l}{\emph{Experiment 3 (deployment-time arbiter):}} \\
    $\pi_{0.5}$ (2B+) & reflex-data policy + arbiter & none & policy + arbiter \\
    \midrule
    \multicolumn{4}{l}{\emph{Appendix-only explicit tactile diagnostics:}} \\
    ACT (18M) & vision+tactile-image policy & left, right tactile images & policy-only \\
    $\pi_{0.5}$ (2B+) & vision+tactile-image policy & left, right tactile images & policy-only \\
    \bottomrule
  \end{tabular}
\end{table}

\subsection{Comparison Design and Evaluation Metrics}
\label{app:exp_design}

\subsubsection{Matched-Comparison Controls}
\label{app:exp_controls}

\paragraph{Matched-comparison controls and alternative explanations.}
All main comparisons are paired to isolate one variable at a time. One
expert operator, who was highly familiar with the robot and collection
system, collected the data on the same platform and task. In both
collection modes, the operator had on-site visual access and saw live
tactile parameters and slip events in the collection GUI. In manual
mode these visually displayed signals were advisory only: gripper
commands still passed through human observation, reaction, and coarse
button execution, without haptic or controller actuation. Only the
reflex condition closed a direct 25\,Hz tactile--action loop.

The original \emph{visually screened manual-data baseline} is not an
unfiltered teleoperation pool. We first recorded 100 manual trials,
then screened the recorded camera streams and retained trials with
successful task completion and no visually observable cup deformation
or obvious grasp instability. From this pool, we selected 30
demonstrations to match the reflex-data count; ACT and $\pi_{0.5}$ use
this same pool in their main matched comparisons. The original manual
trials did not retain tactile traces, so they cannot be screened
post-hoc by CQ or teacher-nearness. Reflex tactile traces were logged
for analysis, not for selecting its 30 demonstrations. We therefore
interpret the main comparison as reflex-regulated collection versus a
visually screened best-effort manual baseline, not as a symmetric
contact-screened comparison. The visually screened manual-data +
arbiter rows reuse the same policy and change only runtime gripper
authority, so they are bridge ablations rather than separate training
baselines. Experiment~2 changes only the training objective within
reflex-data training; Experiment~3 changes only runtime gripper control
for the same reflex-data policy and disturbance protocol.

Reflex data were collected before the manual pools, and collection order
was not randomized. Practice would tend to favor the later manual data,
but temporal effects remain possible.

\subsubsection{Contact-Screened Controlled Ablation}
\label{app:contact_screened}

To control more directly for contact quality, we collected 100 fresh
manual demonstrations with tactile logging and ranked them by
teacher-referenced CQ-near occupancy. We retained the top 30 and
retrained ACT with the same policy inputs, architecture, training
recipe, and deployment protocol. The tactile logs were used only for
offline ranking: they neither actuated the manual gripper nor entered
ACT training or inference. Selecting 30 of 100 is not a collection-yield
estimate.

\begin{table}[h]
  \centering
  \caption{Single-run ACT contact-screened controlled ablation. The
  contact-screened manual set is the top 30 of 100 fresh tactile-logged
  manual demonstrations under offline CQ-near ranking.}
  \label{tab:contact_screened_ablation}
  \footnotesize
  \setlength{\tabcolsep}{3pt}
  \begin{tabular}{lcccc}
    \toprule
    \textbf{Training data} & \textbf{Stable} & \textbf{CQ eps.}
      & \textbf{$\geq$10f run} & \textbf{CQ near teacher} \\
    \midrule
    Reflex-data & \textbf{19/20} & \textbf{19/20} & \textbf{15/20} & \textbf{68.7\%} \\
    Contact-screened manual-data & 6/20 & 5/20 & 5/20 & 24.6\% \\
    \bottomrule
  \end{tabular}
\end{table}

Stable outcomes differ significantly (95\% vs.\ 30\%; two-sided Fisher
exact $p=3.9\times10^{-5}$). The contact-screened condition has one
training and evaluation run and no additional-seed study, so it is
controlled corroboration rather than a multi-seed robustness result.

\subsubsection{Metric Roles and Failure Taxonomy}
\label{app:exp_metrics}

\paragraph{Metric roles at a glance.}
\texttt{Stable}/\texttt{loose}/\texttt{drop} are whole-episode task
outcomes assigned post-trial by consensus among five members of the
author team from wrist- and front-camera videos. \texttt{Stable} means
completed placement followed by secure retention, without visible
migration or irreversible deformation; \texttt{loose} means the object
remains retained but shows visible loosening, migration, deformation,
or unstable placement; and \texttt{drop} means object loss. These are
video outcomes, not force ground truth.
These summaries answer different questions. \emph{CQ episodes}
records whether an episode enters the paper's
conservative tactile contact predicate at least once; it is a
tactile-side regime check and need not coincide with visually stable
task completion. The \emph{$\geq$10-frame teacher-near run} is the
paper's event-style summary of regime entry and sustained re-entry.
\emph{CQ frames near teacher} is the full-horizon duration-weighted
occupancy statistic over contact-qualified frames.
\emph{CQ-onset 3\,s near/CQ occupancy} is used only in Experiment~2 to
measure early teacher-near entry after contact begins, while
\emph{time to release} and \emph{release-time reduction} are
Experiment~2 release-efficiency descriptors rather than replacement
tactile metrics.
For this Experiment~2 onset-window metric, let $\tau_e$ denote the
first timestamp in episode $e$ for which $C_t=1$, where $C_t$
indicates that frame $t$ is contact-qualified and $N_t$ indicates that
frame $t$ is both contact-qualified and teacher-near under the same
stable-hold reference band used throughout the paper; over the window
$[\tau_e,\tau_e+3\,\mathrm{s})$, \emph{CQ-onset 3\,s near/CQ occupancy}
applies the same $\sum_t N_t / \sum_t C_t$ ratio only within that
window. We prefer this onset-window summary to full-horizon occupancy
in Experiment~2 because the refinement is meant to sharpen early
teacher-near entry after contact begins. When release also becomes more
efficient, later teacher-near dwell time can shorten, so a full-horizon
duration-weighted occupancy may decrease even though early regime entry
improves. \emph{Time to release} measures task start to release onset;
\emph{release-time reduction} is relative to the matched reflex-data
baseline.

\paragraph{Qualitative failure taxonomy.}
Across the current rollouts, four recurring failure modes explain most
of the mismatch between task labels and tactile summaries. First,
\emph{no meaningful contact}: the policy never enters CQ. Second,
\emph{unstable hold}: CQ is entered, but sustained teacher-near runs or
high occupancy do not materialize and the cup remains visibly loose.
Third, \emph{delayed release / place hesitation}: the grasp is retained
but release lags during place, which can increase full-horizon CQ
occupancy without implying better early regime entry. Fourth,
\emph{disturbance-induced drop}: the nominal grasp is formed, but a
lateral perturbation during place exceeds the policy-only within-chunk
reactive capacity. This taxonomy is interpretive rather than a
replacement label set, but it clarifies why different summaries can
disagree without contradicting one another.

\subsubsection{Statistical Notes}
\label{app:exp_stats}

\paragraph{Statistical note for the main binary results.}
Table~\ref{tab:binary_stats} reports exact tests and Wilson confidence
intervals for the same single-run binary outcomes already summarized in
the main paper.

\begin{table}[h]
  \centering
  \caption{Key binary comparisons already reported in the main text.
  Intervals are 95\% Wilson confidence intervals; $p$-values are
  two-sided Fisher exact tests.}
  \label{tab:binary_stats}
  \scriptsize
  \setlength{\tabcolsep}{3pt}
  \renewcommand{\arraystretch}{1.03}
  \begin{tabular}{llllc}
    \toprule
    \textbf{Comparison} & \textbf{Endpoint} & \textbf{Condition A}
      & \textbf{Condition B} & \textbf{Fisher $p$} \\
    \midrule
    ACT reflex vs.\ visually screened manual-data & Stable
      & 95.0\% [76.4, 99.1] & 15.0\% [5.2, 36.0]
      & $4.1 \times 10^{-7}$ \\
    ACT reflex vs.\ visually screened manual-data + arbiter & Stable
      & 95.0\% [76.4, 99.1] & 50.0\% [29.9, 70.1]
      & $3.3 \times 10^{-3}$ \\
    ACT reflex vs.\ contact-screened manual-data & Stable
      & 95.0\% [76.4, 99.1] & 30.0\% [14.5, 51.9]
      & $3.9 \times 10^{-5}$ \\
    $\pi_{0.5}$ reflex vs.\ visually screened manual-data & Stable
      & 95.0\% [76.4, 99.1] & 5.0\% [0.9, 23.6]
      & $5.8 \times 10^{-9}$ \\
    $\pi_{0.5}$ reflex vs.\ visually screened manual-data + arbiter & Stable
      & 95.0\% [76.4, 99.1] & 55.0\% [34.2, 74.2]
      & $8.4 \times 10^{-3}$ \\
    Policy-only vs.\ arbiter & Failure
      & 45.0\% [25.8, 65.8] & 0.0\% [0.0, 16.1]
      & $1.23 \times 10^{-3}$ \\
    \bottomrule
  \end{tabular}
\end{table}

\subsection{Additional-Seed Robustness Tables}
\label{app:multiseed}

\paragraph{Protocol.}
The main-text tables report one representative single-run evaluation:
the original run with training seed~42. For the later robustness checks
summarized briefly in the main text, each training variant was retrained
with four additional seeds
($3407, 2025, 7, 123$) under the same dataset, training recipe, and
evaluation protocol; only the training random seed changed.
Experiment~1a, Experiment~1b, and Experiment~3 use 20 rollouts per
additional seed, and Experiment~2 reports seed-wise summary
statistics for the same evaluation quantities used in the main paper.
The following tables retain the full seed-wise values for the reflex
and visually screened manual-data studies. The fresh contact-screened
ablation (Appendix~\ref{app:contact_screened}) has no additional seeds
and is excluded.

\begin{table}[h]
  \centering
  \caption{Experiment~1a additional-seed ACT robustness summary.}
  \label{tab:multiseed_act}
  \scriptsize
  \setlength{\tabcolsep}{2.4pt}
  \renewcommand{\arraystretch}{1.03}
  \resizebox{\linewidth}{!}{%
  \begin{tabular}{r l l c c c c c c}
    \toprule
    \textbf{Seed} & \textbf{ACT variant} & \textbf{Runtime}
      & \textbf{Stable} & \textbf{Loose} & \textbf{Drop}
      & \textbf{CQ eps.} & \textbf{$\geq$10f run}
      & \textbf{CQ near teacher} \\
    \midrule
    3407 & Reflex-data & policy & 19/20 & 1/20 & 0/20 & 19/20 & 15/20 & 67.9\% \\
    2025 & Reflex-data & policy & 20/20 & 0/20 & 0/20 & 20/20 & 16/20 & 72.1\% \\
    7 & Reflex-data & policy & 18/20 & 2/20 & 0/20 & 18/20 & 14/20 & 65.8\% \\
    123 & Reflex-data & policy & 19/20 & 1/20 & 0/20 & 19/20 & 15/20 & 69.0\% \\
    3407 & Visually screened manual-data & policy & 3/20 & 12/20 & 5/20 & 1/20 & 0/20 & 10.2\% \\
    2025 & Visually screened manual-data & policy & 2/20 & 12/20 & 6/20 & 0/20 & 0/20 & 0.0\% \\
    7 & Visually screened manual-data & policy & 4/20 & 9/20 & 7/20 & 1/20 & 0/20 & 13.8\% \\
    123 & Visually screened manual-data & policy & 4/20 & 10/20 & 6/20 & 2/20 & 0/20 & 11.4\% \\
    3407 & Visually screened manual-data & arbiter & 10/20 & 4/20 & 6/20 & 9/20 & 6/20 & 37.8\% \\
    2025 & Visually screened manual-data & arbiter & 9/20 & 3/20 & 7/20 & 11/20 & 8/20 & 41.1\% \\
    7 & Visually screened manual-data & arbiter & 10/20 & 5/20 & 5/20 & 10/20 & 7/20 & 39.0\% \\
    123 & Visually screened manual-data & arbiter & 10/20 & 4/20 & 6/20 & 10/20 & 8/20 & 39.7\% \\
    \bottomrule
  \end{tabular}%
  }
\end{table}

\begin{table}[h]
  \centering
  \caption{Experiment~1b additional-seed $\pi_{0.5}$ ID robustness
  summary.}
  \label{tab:multiseed_pi}
  \scriptsize
  \setlength{\tabcolsep}{2.4pt}
  \renewcommand{\arraystretch}{1.03}
  \resizebox{\linewidth}{!}{%
  \begin{tabular}{r l l c c c c c c}
    \toprule
    \textbf{Seed} & \textbf{$\pi_{0.5}$ variant} & \textbf{Runtime}
      & \textbf{Stable} & \textbf{Loose} & \textbf{Drop}
      & \textbf{CQ eps.} & \textbf{$\geq$10f run}
      & \textbf{CQ near teacher} \\
    \midrule
    3407 & Reflex-data & policy & 18/20 & 2/20 & 0/20 & 17/20 & 11/20 & 67.3\% \\
    2025 & Reflex-data & policy & 20/20 & 0/20 & 0/20 & 19/20 & 13/20 & 72.1\% \\
    7 & Reflex-data & policy & 20/20 & 0/20 & 0/20 & 18/20 & 12/20 & 69.4\% \\
    123 & Reflex-data & policy & 19/20 & 1/20 & 0/20 & 18/20 & 12/20 & 71.0\% \\
    3407 & Visually screened manual-data & policy & 2/20 & 14/20 & 4/20 & 2/20 & 0/20 & 11.2\% \\
    2025 & Visually screened manual-data & policy & 1/20 & 13/20 & 6/20 & 1/20 & 0/20 & 0.0\% \\
    7 & Visually screened manual-data & policy & 0/20 & 14/20 & 6/20 & 0/20 & 0/20 & 0.0\% \\
    123 & Visually screened manual-data & policy & 2/20 & 13/20 & 5/20 & 2/20 & 0/20 & 7.4\% \\
    3407 & Visually screened manual-data & arbiter & 12/20 & 3/20 & 5/20 & 11/20 & 8/20 & 57.8\% \\
    2025 & Visually screened manual-data & arbiter & 10/20 & 6/20 & 4/20 & 9/20 & 6/20 & 51.4\% \\
    7 & Visually screened manual-data & arbiter & 11/20 & 4/20 & 5/20 & 11/20 & 8/20 & 58.6\% \\
    123 & Visually screened manual-data & arbiter & 9/20 & 7/20 & 4/20 & 8/20 & 6/20 & 53.2\% \\
    \bottomrule
  \end{tabular}%
  }
\end{table}

\begin{table}[h]
  \centering
  \caption{Experiment~2 additional-seed refinement summary.}
  \label{tab:multiseed_dyn}
  \scriptsize
  \setlength{\tabcolsep}{2.8pt}
  \renewcommand{\arraystretch}{1.03}
  \resizebox{\linewidth}{!}{%
  \begin{tabular}{r l l c c c c}
    \toprule
    \textbf{Seed} & \textbf{Backbone} & \textbf{Variant}
      & \textbf{$\geq$10f run}
      & \shortstack{\textbf{CQ-onset 3\,s}\\\textbf{near/CQ}}
      & \shortstack{\textbf{Time to}\\\textbf{release}}
      & \shortstack{\textbf{Release-time}\\\textbf{reduction}} \\
    \midrule
    3407 & ACT & baseline & 14/20 & 70.4\% & 16.1\,s & --- \\
    2025 & ACT & baseline & 16/20 & 74.1\% & 15.0\,s & --- \\
    7 & ACT & baseline & 15/20 & 71.8\% & 16.4\,s & --- \\
    123 & ACT & baseline & 15/20 & 73.0\% & 15.4\,s & --- \\
    3407 & ACT & + dyn.\ loss & 17/20 & 81.0\% & 13.8\,s & 14.3\% \\
    2025 & ACT & + dyn.\ loss & 18/20 & 84.3\% & 12.9\,s & 14.0\% \\
    7 & ACT & + dyn.\ loss & 18/20 & 82.1\% & 13.2\,s & 19.5\% \\
    123 & ACT & + dyn.\ loss & 18/20 & 83.0\% & 13.5\,s & 12.3\% \\
    3407 & $\pi_{0.5}$ & baseline & 10/20 & 77.4\% & 29.3\,s & --- \\
    2025 & $\pi_{0.5}$ & baseline & 12/20 & 79.1\% & 27.9\,s & --- \\
    7 & $\pi_{0.5}$ & baseline & 12/20 & 78.0\% & 28.7\,s & --- \\
    123 & $\pi_{0.5}$ & baseline & 13/20 & 79.2\% & 27.4\,s & --- \\
    3407 & $\pi_{0.5}$ & + dyn.\ loss & 15/20 & 84.0\% & 22.4\,s & 23.5\% \\
    2025 & $\pi_{0.5}$ & + dyn.\ loss & 16/20 & 85.8\% & 22.0\,s & 21.1\% \\
    7 & $\pi_{0.5}$ & + dyn.\ loss & 16/20 & 85.0\% & 21.5\,s & 25.1\% \\
    123 & $\pi_{0.5}$ & + dyn.\ loss & 17/20 & 87.6\% & 20.9\,s & 23.7\% \\
    \bottomrule
  \end{tabular}%
  }
\end{table}

\paragraph{Experiment~3 disturbance illustration and additional-seed summary.}
Fig.~\ref{fig:exp3_disturbance} shows two example disturbance
configurations from the randomized external-disturbance evaluation, and
Table~\ref{tab:multiseed_exp3} reports the matching additional-seed
runtime robustness summary.

\begin{center}
  \begin{minipage}{\linewidth}
  \centering
  \begin{minipage}[t]{0.56\linewidth}
    \vspace{0pt}
    \centering
    \makebox[\linewidth][c]{%
      \begin{minipage}[t]{0.485\linewidth}
        \centering
        \includegraphics[width=\linewidth]{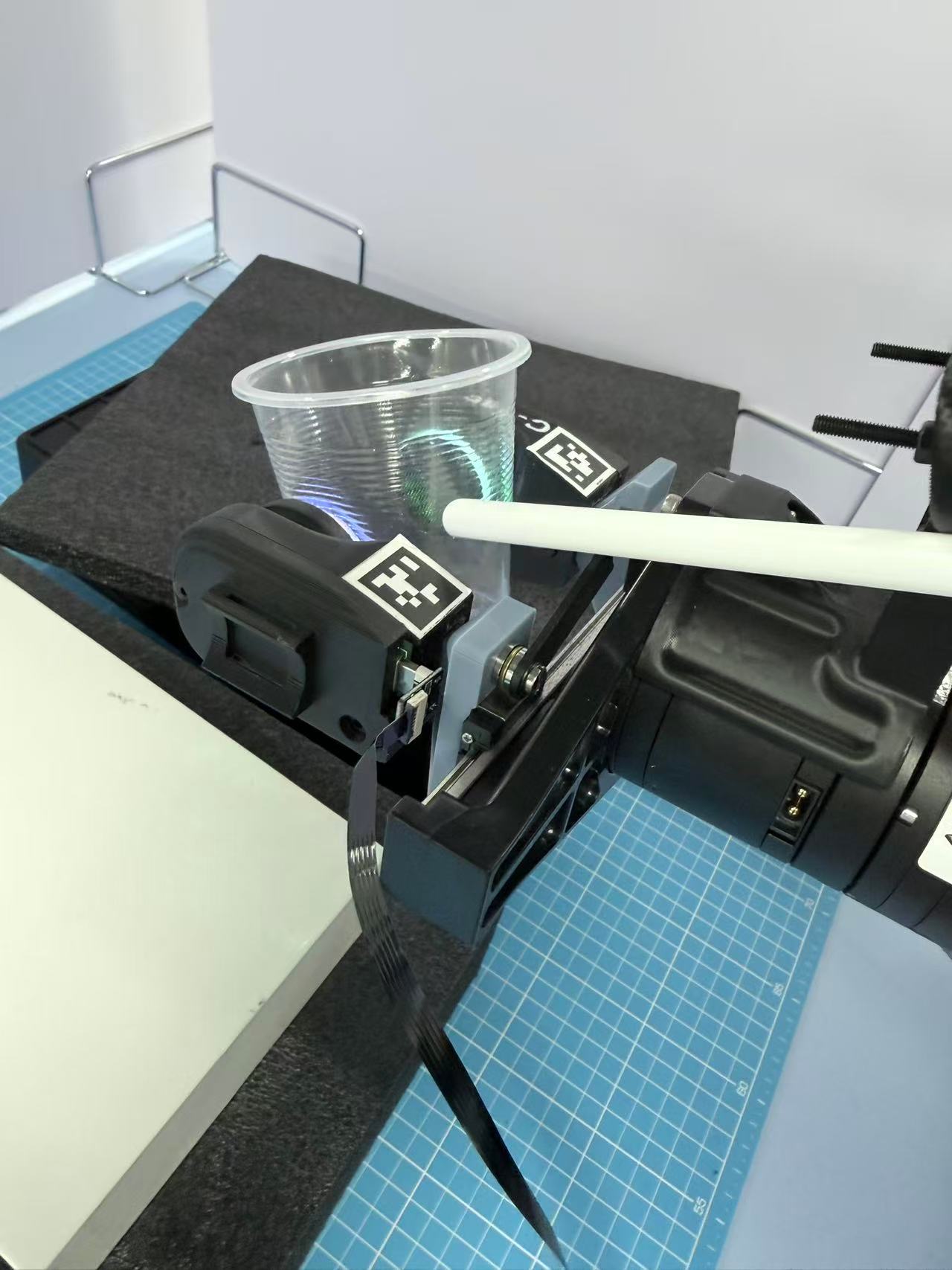}
      \end{minipage}\hfill
      \begin{minipage}[t]{0.485\linewidth}
        \centering
        \includegraphics[width=\linewidth]{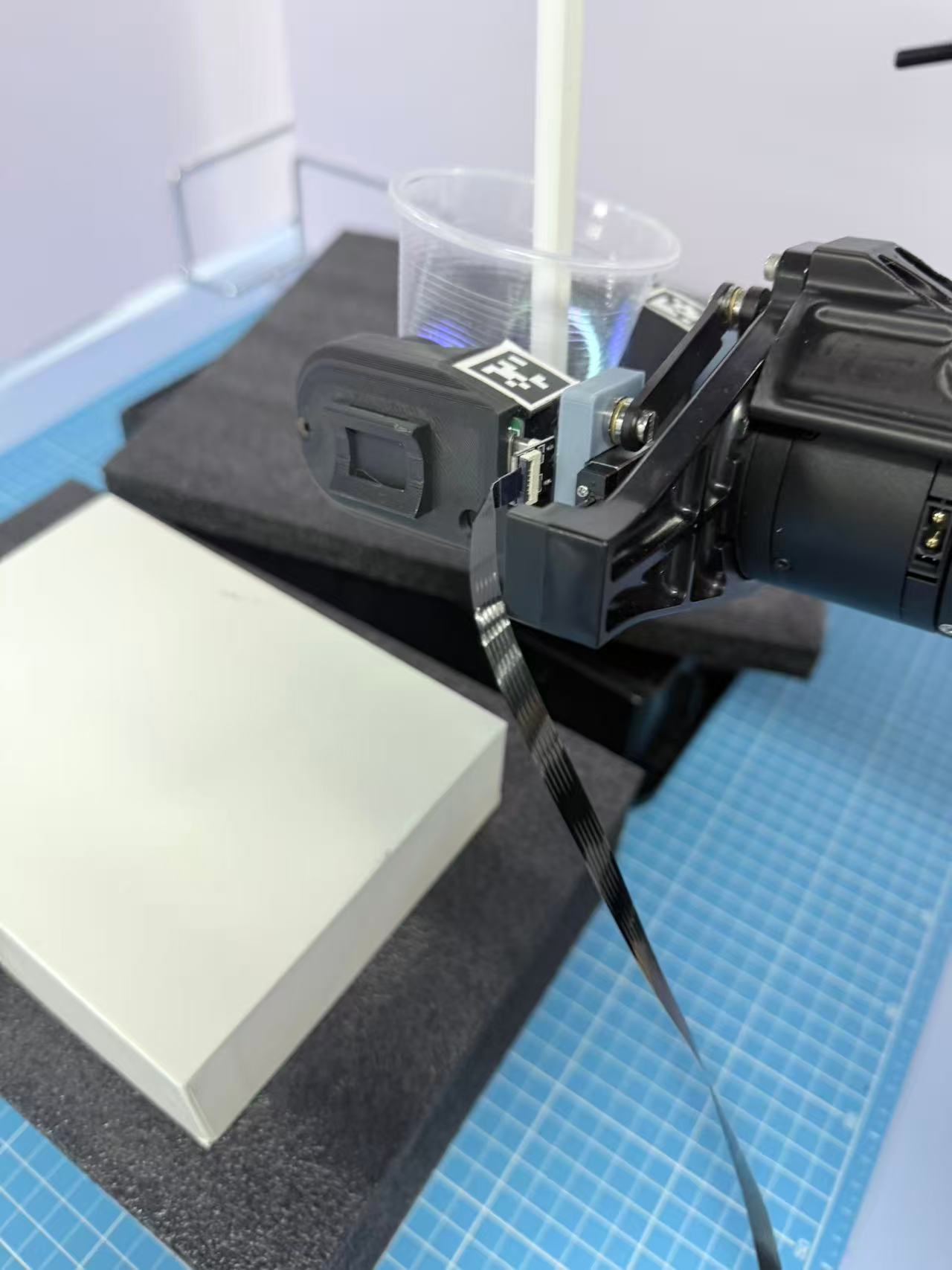}
      \end{minipage}%
    }
    \par\vspace{3pt}
    \captionof{figure}{Experiment~3 disturbance configurations. Two
    representative stick-contact perturbations are shown after stable
    lift during the place phase.}
    \label{fig:exp3_disturbance}
  \end{minipage}\hfill
  \begin{minipage}[t]{0.40\linewidth}
    \vspace{0pt}
    \captionof{table}{Experiment~3 additional-seed runtime robustness
    summary.}
    \label{tab:multiseed_exp3}
    \vspace{3pt}
    \centering
    \footnotesize
    \setlength{\tabcolsep}{2pt}
    \renewcommand{\arraystretch}{1.04}
    \begin{tabular*}{\linewidth}{@{\extracolsep{\fill}} r l c c}
      \toprule
      \textbf{Seed} & \textbf{Runtime mode}
        & \textbf{Retained} & \textbf{Failure} \\
      \midrule
      3407 & policy-only & 10/20 & 10/20 \\
      2025 & policy-only & 12/20 & 8/20 \\
      7 & policy-only & 11/20 & 9/20 \\
      123 & policy-only & 9/20 & 11/20 \\
      3407 & policy + arbiter & 20/20 & 0/20 \\
      2025 & policy + arbiter & 19/20 & 1/20 \\
      7 & policy + arbiter & 20/20 & 0/20 \\
      123 & policy + arbiter & 20/20 & 0/20 \\
      \bottomrule
    \end{tabular*}
  \end{minipage}
  \end{minipage}
\end{center}

\subsection{Supplementary Experiment 1 Details}
\label{app:exp_exp1}

\subsubsection{$\pi_{0.5}$ Tactile Corroboration}
\label{app:exp_pi05_tactile}

\begin{figure}[h]
  \centering
  \begin{minipage}[c]{0.62\linewidth}
    \centering
    {\footnotesize \textbf{(a)} Teacher-referenced tactile force distribution.\par}
    \includegraphics[width=0.8\linewidth]{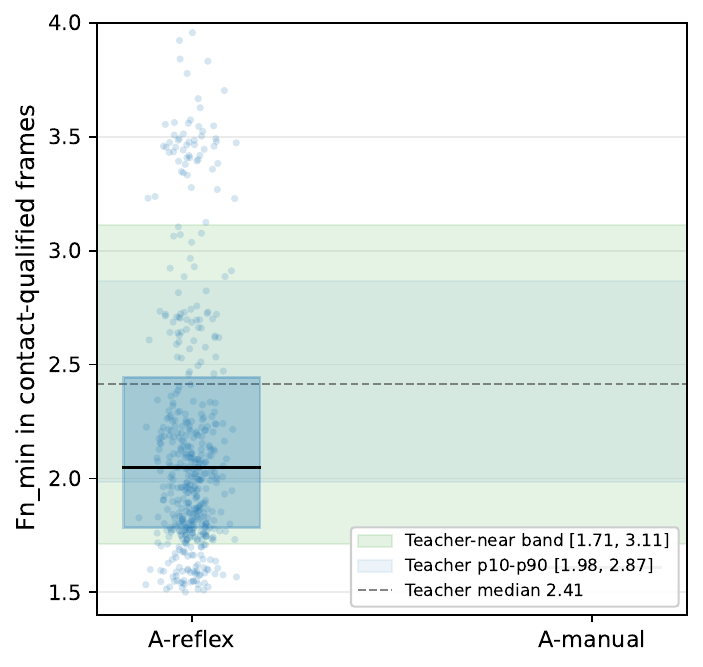}
  \end{minipage}\hfill
  \begin{minipage}[c]{0.34\linewidth}
    \centering
    {\footnotesize \textbf{(b)} Numerical summary.\par}
    \scriptsize
    \setlength{\tabcolsep}{2pt}
    \renewcommand{\arraystretch}{1.03}
    \begin{tabular}{lccc}
      \toprule
      \shortstack{\textbf{$\pi_{0.5}$}\\\textbf{policy}} &
      \shortstack{\textbf{CQ}\\\textbf{eps.}} &
      \shortstack{\textbf{$\geq$10f}\\\textbf{run}} &
      \shortstack{\textbf{CQ near}\\\textbf{teacher}} \\
      \midrule
      Reflex-data & \textbf{18/20} & \textbf{12/20} & \textbf{70.8\%} \\
      \shortstack{visually screened\\manual-data} & 1/20 & 0/20 & 0.0\% \\
      \shortstack{visually screened\\manual-data\\+ arbiter} & 11/20 & 8/20 & 56.5\% \\
      \bottomrule
    \end{tabular}
  \end{minipage}
  \caption{$\pi_{0.5}$ teacher-referenced tactile corroboration for the
  ID comparison. (a) Policy-only
  comparison against the teacher reference. (b) Compact summary of the
  same diagnostic,
  including the visually screened manual-data policy + arbiter bridge
  ablation.}
  \label{fig:pi05_teacher_hold}
\end{figure}

\paragraph{$\pi_{0.5}$ teacher-referenced tactile corroboration.}
Fig.~\ref{fig:pi05_teacher_hold} reports the appendix version of the
$\pi_{0.5}$ teacher-referenced tactile corroboration used to support
Experiment~1b in the main text.

The next two notes provide appendix-only scope details for the
paper-cup OOD evaluation and the sparse-CQ ACT cases.

\subsubsection{Paper-Cup Scope Notes}
\label{app:exp_papercup}

\begin{figure}[h]
  \centering
  \includegraphics[width=0.40\linewidth]{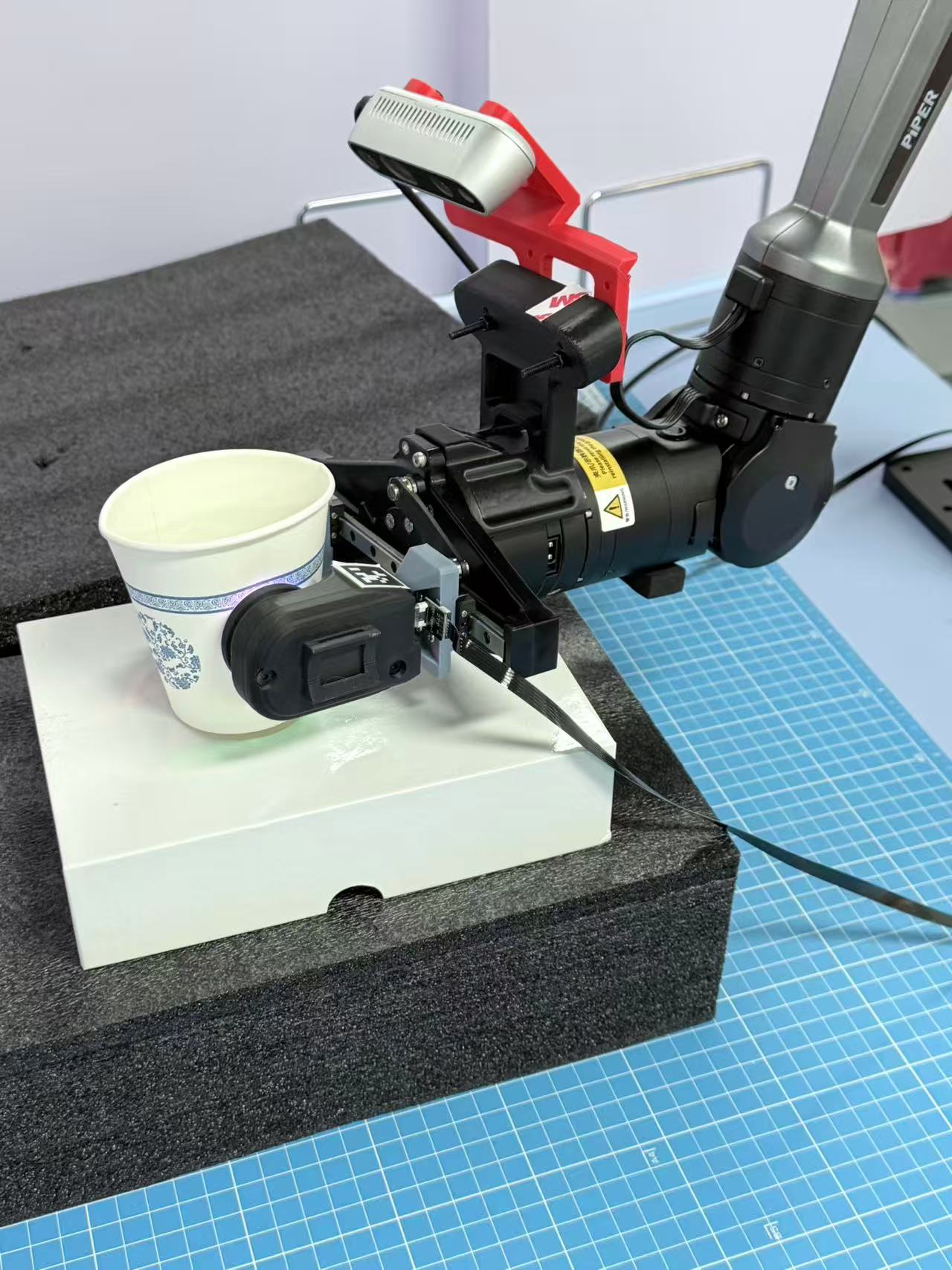}
  \caption{Unseen paper-cup OOD setup used in Experiment~1c.
  The image shows the direct-contact grasping configuration for the
  unseen paper cup used in the OOD evaluation. Compared with
  the training plastic cup, this object has a larger force margin,
  higher friction, and clearer visual boundaries, so we treat it as an
  OOD setting rather than as a second high-power confirmatory
  benchmark.}
  \label{fig:paper_cup_setup}
\end{figure}

\paragraph{Unseen paper-cup OOD setup.}
Fig.~\ref{fig:paper_cup_setup} shows the direct-contact grasping
setup used in the unseen paper-cup OOD evaluation from
Experiment~1c.

\paragraph{Unseen paper-cup OOD statistical note.}
For the paper-cup extension in Experiment~1c, the reflex-data policy
achieves 80.0\% stable [49.0, 94.3] versus 30.0\% [10.8, 60.3] for the
visually screened manual-data policy. The numerical difference is not
statistically significant (two-sided Fisher exact $p=0.0698$), so we
present it only as exploratory OOD scope evidence rather than as a
second confirmatory claim.

\subsubsection{Sparse-CQ and ACT Boundary Notes}
\label{app:exp_sparsecq}

\paragraph{Sparse-CQ caveats in ACT tactile diagnostics.}
The $12.5\%$ value for the visually screened manual-data ACT policy in
Table~\ref{tab:exp1a_act_tactile} comes from a single
contact-qualified episode containing only eight CQ frames, so it does
not correspond to any sustained teacher-near run. Because
\emph{CQ frames near teacher} is conditioned on CQ frames only, the
$39.4\%$ value for the visually screened manual-data ACT policy + arbiter
should also be read conditionally: it is a CQ-conditioned occupancy
statistic, not a whole-episode success fraction. In the updated ACT
arbiter evaluation, the same row corresponds to 10/20 CQ episodes and
7/20 sustained teacher-near runs, so we read it as partial recovery
rather than regime-level parity with the reflex-data policy.

\paragraph{ACT paper-cup boundary note.}
For completeness, the reflex-data ACT policy can still grasp the unseen
paper cup, but it often stalls after grasp. We therefore keep
Experiment~1a as the primary controlled comparison and use $\pi_{0.5}$ for
the main text's OOD evidence.

\section{Dynamics-Loss Refinement Notes}
\label{app:dyn_refinement}

This appendix records supporting notes for Experiment~2. The
dynamics-loss objective itself is defined in
Sec.~\ref{sec:method:secondary_refinement}, and the metric definitions
are given in Appendix~\ref{app:exp_metrics}. Here we keep only the
interpretive notes needed to explain the omitted full-horizon
diagnostic and the choice of the CQ-onset 3\,s summary.

\paragraph{Main-text summary and omitted diagnostic.}
Table~\ref{tab:exp2_dyn} in the main text reports the matched
comparisons most directly aligned with the refinement's intended
benefit: sustained teacher-near entry,
\emph{CQ-onset 3\,s near/CQ occupancy}, \emph{time to release}, and
\emph{release-time reduction}. We emphasize the CQ-onset 3\,s summary
because Experiment~2 asks whether the refinement sharpens early
teacher-near entry after meaningful contact begins. A full-horizon
duration-weighted occupancy can move in the opposite direction when
release becomes more efficient: the baseline may simply accumulate more
later teacher-near dwell time by holding longer, whereas the
dynamics-loss policy can enter earlier but also release sooner and
thus spend fewer later contact-qualified frames near the teacher. In
other words, earlier entry and earlier release onset can improve the onset-side
summary while reducing the total time later spent stably dwelling near
the teacher, so a full-horizon occupancy statistic can decrease even
when the contact transition itself becomes sharper.
For completeness, the omitted full-horizon \emph{CQ near teacher}
occupancy decreases from 68.7\% to 54.8\% on ACT and from 70.8\% to
61.2\% on $\pi_{0.5}$ when the dynamics loss is added. We treat that
statistic as a caveat rather than as the primary summary of the
refinement.

\paragraph{Interpretation and caveat.}
Across both backbones, we read the refinement narrowly: it is
most consistent with stronger early regime entry and shorter time to
release inside the controller-shaped regime, not with uniformly higher
values on every duration-weighted tactile summary. We treat it as a
scoped train-time refinement rather than a co-equal contribution,
because the gains are not uniform across summaries or task outcomes,
and we have not yet tested the same loss under manual-data training, on
additional backbones, or under a fully unified task-level outcome
protocol.

\end{document}